\documentclass{article}

\usepackage{arxiv}

\usepackage[utf8]{inputenc} 
\usepackage[T1]{fontenc}    
\usepackage{hyperref}       
\usepackage{url}            
\usepackage{booktabs}       
\usepackage{amsfonts}       
\usepackage{nicefrac}       
\usepackage{microtype}      
\usepackage{lipsum}		
\usepackage{graphicx}
\usepackage{natbib}
\usepackage{doi}
\usepackage{xcolor}
\usepackage{tikz}
\makeatletter
\providecommand{\tikz@picmode}{}
\makeatother
\usetikzlibrary{positioning,fit,arrows.meta,patterns,calc}
\usetikzlibrary{decorations.pathreplacing,calligraphy} 

\usepackage{amsmath}
\usepackage{pgfplots}
\pgfplotsset{compat=1.18}
\usetikzlibrary{arrows.meta}
\usepackage{pstricks}
\usepackage[absolute,overlay]{textpos}

\definecolor{tokcolor}{RGB}{130,130,130}
\definecolor{bytecolor}{RGB}{255,255,0}
\definecolor{sumcolor}{RGB}{80,160,100}

\title{Spelling Bee Embeddings for Language Modeling}

\author{
Markus N. Rabe \\
Sutter Hill Ventures
\AND
Judith Clymo \\
University of California, Santa Cruz
\AND
Zheren Dong \\
Independent Researcher
}

\renewcommand{\shorttitle}{Spelling Bee Embeddings}

\begin{document}

\maketitle

\begin{abstract}
We introduce a simple modification to the embedding layer.
The key change is to infuse token embeddings with information about their spelling.
Models trained with these embeddings improve not only on spelling tasks, but also across standard benchmarks.
We conduct scaling studies for models with 40M to 800M parameters, which suggest the improvements are equivalent to needing about 8\% less compute and data to achieve the same test loss.
The improvements also translate to gains on a wide set of benchmarks.
\end{abstract}


\section{Introduction}
\label{sec:introduction}
The question to ``count the number of Rs in strawberry'' is a popular example that demonstrates the unintuitive failure modes of language models.
It is a seemingly trivial question that tripped up even the largest language models, which insisted that strawberry contains just two Rs.
The failure is widely attributed to tokenization, i.e., the process of grouping characters in a text into tokens before applying a language model.

To develop an intuition about tokenization challenges, let's consider how GPT-4 tokenizes variants of the word strawberry:

\vspace{-.4cm}
\newcommand{\tbox}[2][gray!20]{%
  \fcolorbox{black}{#1}{\strut #2}%
}
\quad\begin{minipage}[t]{0.4\textwidth}

\smallskip
\verb|strawberry| \hfill \tbox[blue!15]{str}\tbox[yellow!15]{aw}\tbox[red!15]{berry}

\verb|␣strawberry| \hfill \tbox[magenta!15]{ strawberry}

\verb|Strawberry| \hfill \tbox[brown!15]{Str}\tbox[yellow!15]{aw}\tbox[red!15]{berry}

\verb|␣Strawberry| \hfill \tbox[yellow!35]{ Strawberry}

\verb|(strawberry| \hfill \tbox[gray!45]{(str}\tbox[yellow!15]{aw}\tbox[red!15]{berry}

\verb|(Strawberry| \hfill \tbox[green!15]{(Str}\tbox[yellow!15]{aw}\tbox[red!15]{berry}

\verb|,strawberry| \hfill \tbox[olive!15]{,str}\tbox[yellow!15]{aw}\tbox[red!15]{berry}

\verb|=strawberry| \hfill \tbox[orange!60]{=str}\tbox[yellow!15]{aw}\tbox[red!15]{berry}

\end{minipage}
\qquad\qquad\qquad
\begin{minipage}[t]{0.4\textwidth}

\smallskip


\verb|STRAWBERRY| \hfill \tbox[green!35]{ST}\tbox[olive!45]{RAW}\tbox[blue!25]{B}\tbox[red!15]{ERRY}

\verb|'STRAWBERRY| \hfill \tbox[green!35]{'S}\tbox[brown!15]{TR}\tbox[yellow!35]{AW}\tbox[blue!25]{B}\tbox[red!15]{ERRY}

\verb|'strawberry| \hfill \tbox[yellow!60]{'s}\tbox[red!35]{tr}\tbox[yellow!15]{aw}\tbox[red!15]{berry}

\verb|.Strawberry| \hfill \tbox[violet!15]{.Str}\tbox[yellow!15]{aw}\tbox[red!15]{berry}

\verb|.strawberry| \hfill \tbox[violet!45]{.str}\tbox[yellow!15]{aw}\tbox[red!15]{berry}

\verb|[strawberry| \hfill \tbox[cyan!35]{[str}\tbox[yellow!15]{aw}\tbox[red!15]{berry}

\verb|-strawberry| \hfill \tbox[green!40]{-str}\tbox[yellow!15]{aw}\tbox[red!15]{berry}


\verb|_strawberry| \hfill \tbox[blue!60]{\_str}\tbox[yellow!15]{aw}\tbox[red!15]{berry}

\end{minipage}
\vspace{0.2cm}

We observe that tokenization depends heavily on leading spaces, surrounding special characters, and capitalization.
In total, 22 distinct tokens are involved in the variants above.
And these examples above do not even consider plural forms, misspellings, and splitting the word due to line breaks etc.
Any language model must learn to associate all these token sequences with the same underlying word and spelling.

This is not an isolated case. The word "apple" occurs 11 times in GPT4's vocabulary: ` Apple', `apple', ` apple', `Apple', `.apple', ` apples', ` AppleWebKit', `/apple', ` pineapple', `APPLE', ` APPLE'.
These variants need to have different tokenizations, such that the model can distinguish them, but it also means that models need to learn the word's spelling many times over.

Languages with heavier use of inflections suffer even more from tokenization artifacts.
Consider the variants of the German word ``hoch'' (=high): hoch, hohe, hohem, hohen, hoher, hohes, höher, höhere, höherem, höheren, höherer, höheres, höchst, höchste, höchstem, höchsten, höchster, and höchstes.
Their tokenizations make use of 19 distinct tokens---and this does not even account for capitalization, leading spaces or special characters, or compound words.

While language models \emph{can} learn about the spelling of tokens with enough data, we argue that \emph{this makes the learning process less efficient}.
Indeed, even models with hundreds of billions of parameters trained on trillions of tokens struggle to reliably count the number of Rs in strawberry.
While many models have improved on the canonical ‘strawberry’ prompt over time, even small variants trip up recent models.
We present some examples in Appendix~\ref{app:screenshots}.




\begin{figure}[t]
\centering
\begin{tikzpicture}[
    font=\small,
    box/.style={rectangle, draw, rounded corners=2pt, minimum height=0.7em, inner sep=2pt},
    vec/.style={rectangle, draw, rounded corners=2pt, inner sep=4pt, minimum width=3cm},
    >={Latex[length=2mm]},
    node distance=0.8cm and 0.8cm
]

\node[box] (toktext) {\verb|strawberry|};


\node[vec, fill=yellow!50, above right=3.9cm and 1cm of toktext] (byte1) {byte embedding for \texttt{s}};
\node[vec, fill=yellow!50, below=0.1cm of byte1] (byte2) {byte embedding for \texttt{t}};
\node[vec, fill=yellow!50, below=0.1cm of byte2] (byte3) {byte embedding for \texttt{r}};
\node[vec, fill=yellow!50, below=0.1cm of byte3] (byte4) {\hspace{1.2cm}\dots\hspace{0.95cm}\phantom{A}};

\draw[->] (toktext) |- node[above]{\hspace{1.5cm}spelling lookup} (byte1);
\draw[->] (toktext) |- (byte2);
\draw[->] (toktext) |- (byte3);
\draw[->] (toktext) |- (byte4);

\node[vec, fill=yellow!50, right=0.9cm of byte1] (byte1r) {position-aware \texttt{s}};
\node[vec, fill=yellow!50, right=0.9cm of byte2] (byte2r) {position-aware \texttt{t}};
\node[vec, fill=yellow!50, right=0.9cm of byte3] (byte3r) {position-aware \texttt{r}};
\node[vec, fill=yellow!50, right=0.9cm of byte4] (byte4r) {\hspace{1.2cm}\dots\hspace{0.95cm}\phantom{A}};

\draw[->] (byte1) -- node[above]{RoPE} (byte1r);
\draw[->] (byte2) -- (byte2r);
\draw[->] (byte3) -- (byte3r);
\draw[->] (byte4) -- (byte4r);

\draw [decorate, decoration = {brace}] (8.9,1.6) -- node[above] {sum \& scale} (5.7,1.6);


\node[vec, fill=yellow!50, below=0.8cm of byte4r] (charsum) {spelling embedding};

\node[vec, fill=tokcolor!30, below=0.5cm of charsum] (tokemb) {token embedding $e_{\text{tok}}$};
\draw[->] (toktext) -- node[above,pos=0.45]{lookup in embedding table} (tokemb);


\node[
    vec,
    fill=sumcolor!20,
    above right=0.0cm and 2.2cm of tokemb,
    fill=tokcolor!40,        
    path picture={
    \begin{scope}[rounded corners=0pt]
    \foreach \i in {0,...,10} {
      \ifodd\i\relax\else
        \path[pic actions, fill=yellow!50]
          ($(path picture bounding box.south west)!{\i/11}!(path picture bounding box.south east)$)
          rectangle
          ($(path picture bounding box.north west)!{(\i+1)/11}!(path picture bounding box.north east)$);
      \fi
    }
    \end{scope}
    },
] (final) {spelling bee embedding};

\draw[->] (tokemb.east) -- ++(0.8,0) |- (final.west);  
\draw[->] (charsum.east) -- ++(0.8,0) |- node[above]{\hspace{1.3cm}average} (final.west);


\end{tikzpicture}
\caption{Spelling bee embeddings augment a standard token embedding $e_{\text{tok}}$ with a normalized sum of byte-level embeddings for the first 16 bytes of the token. Byte embeddings are position-encoded with RoPE using character positions inside the token.}
\label{fig:spelling-bee-embeddings}
\end{figure}
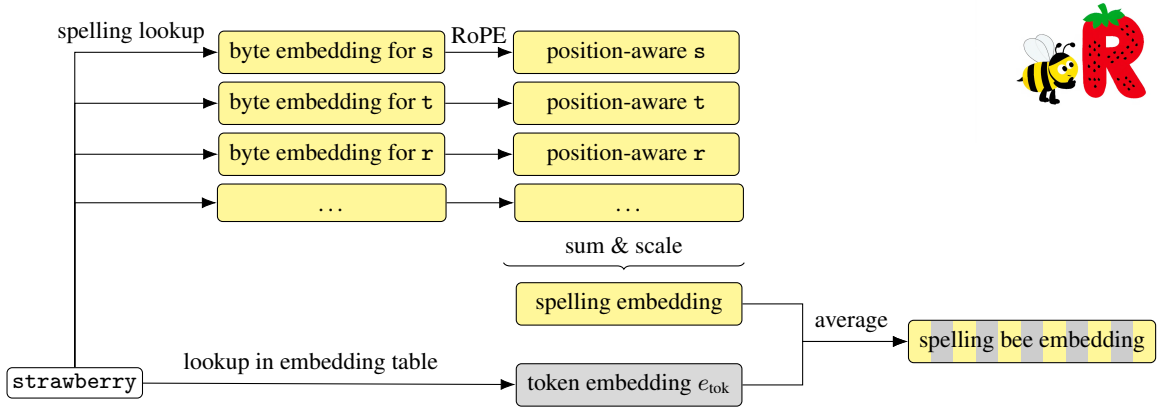

\begin{textblock*}{3cm}(16.5cm,2cm)
    \includegraphics[width=3cm]{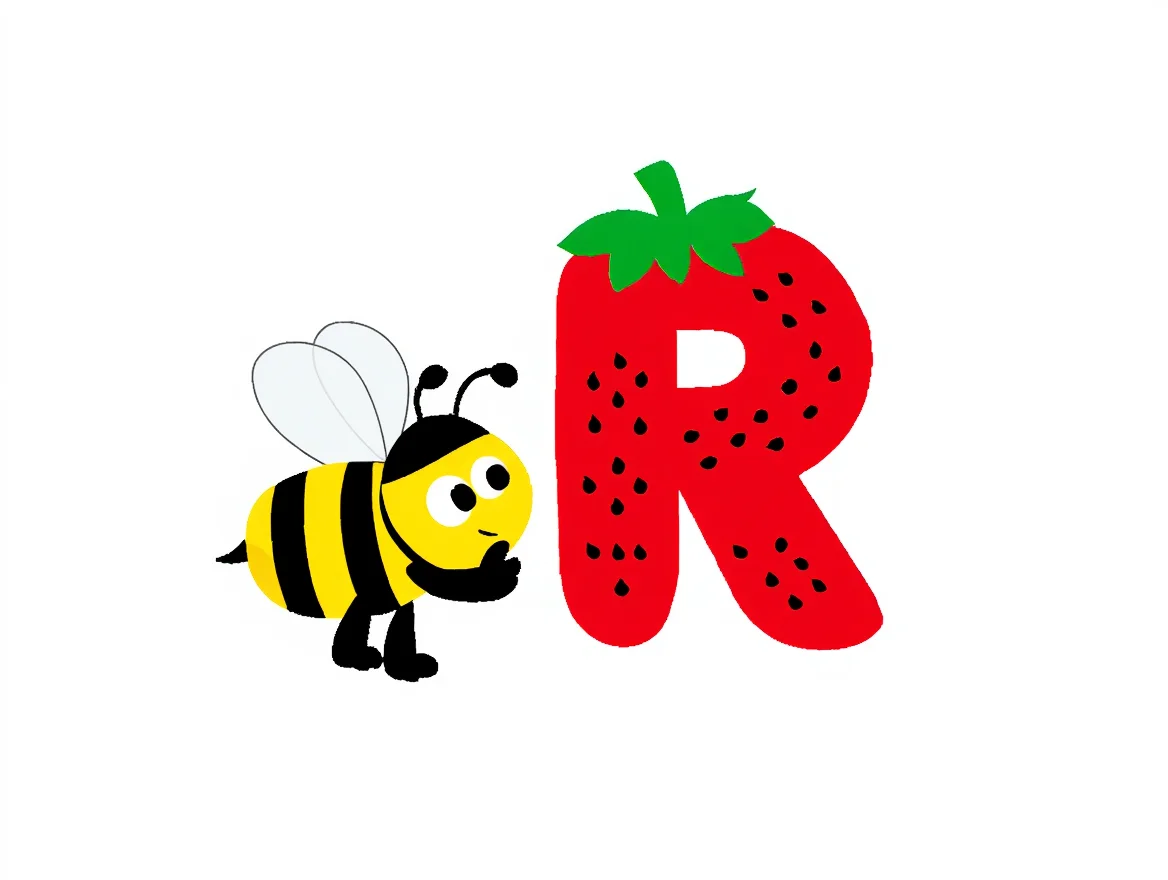}
\end{textblock*}

In this paper, we propose \textbf{spelling bee embeddings}, a simple architectural change that helps language models with the spelling of tokens.
Spelling bee embeddings introduce a negligible amount of additional parameters and essentially no computational overhead.
Compared to a well-tuned baseline, spelling bee embeddings consistently improve the performance of models from 40m parameters to 800m parameters.
We also confirm that models trained with spelling bee embeddings can determine the number of Rs in strawberry.

Like regular token embeddings, spelling bee embeddings have 1 embedding per token, but we also introduce 256 new embeddings, one for each possible value of a byte.
The output of spelling bee embeddings is the addition of the token embedding and the embeddings of the first 16 bytes of the token.
We further improve this "bag of characters" representation by applying rotary position embeddings to the characters, using their position in the token instead of the sequence position.

\section{Related Work}
\label{sec:related}

\citet{shin2024large} compare the performance of language models to humans on spelling challenges, and find that language models indeed struggle on these tasks compared to humans.
\citet{fu2024large} present detailed statistics on spelling challenges in LLMs and suggest that spelling challenges are not related to the frequency of words in the training data.

There have been many attempts to make models aware of character-level structure, e.g. by removing the need for tokenization from language modeling.
Most notably are language models that operate on the level of bytes.
Character-level Transformers (without tokenization) can match or exceed token-based models given the same data.
However, this comes at a dramatically increased computational cost, as the sequence length increases significantly for the same number of text characters.

Some efforts have been made to learn token-like groupings inside the model and thereby keep the computation cost at bay.
Notable examples are Charformer~\citep{tay2021charformer} and MegaByte~\citep{yu2023megabyte}.
However, these methods didn't control for sequence length and compute cost.

\cite{nawrot2023efficient} and \cite{pagnoni2024bytelatenttransformerpatches} proposed variants of transformers that learn to segment the text in an end-to-end fashion.
While their work is beautiful and may eventually obsolete tokenization (and spelling bee embeddings) alltogether, they are also more risky to adopt for existing large training runs due to their complexity.

In contrast, spelling bee embeddings keep the changes to the architecture very local to the embedding layer, which makes them compatible with language models that don't follow the Transformer architecture and greatly simplify their adoption.
Spelling bee embeddings also work well with existing tokenizers, so that existing handling of special tokens etc does not have to change.
This also enables a clean apples-to-apples comparison, as the compute cost and sequence length are unaffected.


\section{Spelling Bee Embeddings}
\label{sec:spellingbee}
The core idea behind \emph{spelling bee embeddings} is to make model embeddings aware of the spelling of tokens, without sacrificing the computational advantages of subword tokenization.
Instead of relying on the model to infer character-level information implicitly, we augment the token embeddings with a compact representation of the token's spelling.

Formally, for each token ID $t$ in the vocabulary, we introduce:
\begin{itemize}
    \item A learned token embedding $e_{\text{tok}}(t) \in \mathbb{R}^d$, as in standard models;
    \item A sequence of at most $16$ bytes $\{b_1, b_2, \dots, b_{16}\}$ obtained from its UTF-8 encoding. If the token's UTF8 encoding is shorter than 16 bytes, we pad with the null byte; if longer, we truncate.
\end{itemize}

Each byte $b_i \in \{0, \dots, 255\}$ is embedded through a learned lookup table of size $256 \times d$, producing byte embeddings $\{e_{b_1}, \dots, e_{b_{16}}\}$. To distinguish the contribution of each character within the token, we apply rotary positional embeddings~\citep{su2024roformer} to the byte embeddings, using the position of each byte inside the token:
\[
\tilde{e}_{b_i} = \text{RoPE}(e_{b_i}, i).
\]

We then compute the combined character embedding for the token:
\[
e_{\text{chars}}(t) = \frac{1}{\alpha}\ \sum_{i=1}^{16} \tilde{e}_{b_i},
\]
where $\alpha$ is a constant chosen such that $\mathbb{E}[\|e_{\text{chars}}(t)\|^2] \approx \mathbb{E}[\|e_{\text{tok}}(t)\|^2]$.

Finally, the full embedding for token $t$, called the \emph{spelling bee embedding} $e_{\text{bee}}$, is simply the mean of $e_{\text{tok}}(t)$ and~$e_{\text{chars}}(t)$.

This construction is illustrated in Figure~\ref{fig:spelling-bee-embeddings}. Importantly, it preserves the original model architecture: all downstream components operate on the same embedding dimension, and the computational and memory overhead is negligible.


\section{Experimental Setup}
\label{sec:setup}

The primary goal of this paper is to evaluate the impact of spelling bee embeddings on the quality of the models.

One of the main challenges with architecture studies is that is that gains often diminish with scale.
To mitigate this risk with acceptable compute cost, we conduct \textbf{scaling studies} where we train models from 44m to 816m parameters.
We establish that the benefit of spelling bee embeddings remains stable as we scale up the models, suggesting that significantly larger models might still benefit from spelling bee embeddings.

Another big challenge in architecture research is that changing the FLOPs requirements means that we need to train models on different amounts of data---which in particular means we train them on different data.
Spelling bee embeddings introduce negligible additional FLOPs, which simplifies doing a strawberries-to-strawberries comparison.
For each model size, we train baseline models and spelling bee models on the exact same data in the same order.

All our experiments train models from random initialization, i.e. we do pretraining.
The code for the final experiments can be found on GitHub: \url{https://github.com/MarkusRabe/littletrainingloop}.

\paragraph{Datasets.}
We use the SlimPajama training set~\citep{cerebras2023slimpajama}.
We pack documents by concatenating text until reaching the sequence length.
Overhanging tokens are carried over to the next batch like in the data pipeline of Memorizing Transformers~\cite{wu2022memorizing}.

For our evaluations, we rely on the \texttt{lm-evaluation-harness}~\citep{eval-harness}.

\paragraph{Architecture.}
Spelling bee embeddings are compatible with a wide range of language model architectures, since they only change the input embeddings.
In this work, however, we focus on the Transformer architecture~\citep{vaswani2017attention}.
As a baseline, we use a Transformer with the SwiGLU activation function~\citep{shazeer2020glu}, rotary position embeddings\citep{su2024roformer}, pre-layer-norm~\citep{baevski2018adaptive, xiong2020layer}, and grouped query attention~\citep{ainslie2023gqa}.
We remove all biases from the linear layers and the layer norms.

\paragraph{Notes on efficiency.}
We use mixed precision training and validated that the results match the FP32 training runs.
This includes the use of the AdamW optimizer from the \texttt{optimi} package that keeps the memory requirements low.
We use FlashAttention~\citet{dao2022flashattention}/memory-efficient attention~\citet{rabe2021self} to keep the memory requirements of attention subquadratic, and we use activation checkpointing to further reduce memory requirements.
For small models, most of the memory is used up by the logit computation.
We split up the logit computation into sections of 4096 along the sequence dimension and use activation checkpointing to reduce the memory footprint.
This allows us to run models beyond 1B parameters with the full batch size on a single H100.

\paragraph{Hyperparameters.}
For our scaling studies we use the family of models of different sizes used also in the Chinchilla paper \citep{hoffmann2022training}.
For the model with "816m" parameters the hyperparameters are as follows:

\begin{tabular}{ll}
    num\_layers & 25 \\
    num\_heads & 12 \\
    head\_dim & 128 \\
    embedding\_dim & 1536
\end{tabular}

We differ from the Chinchilla setting in that we use a larger vocabulary (\texttt{cl100k\_base}, used also by GPT4) with 100k tokens, and grouped query attention~\citep{ainslie2023gqa}.
This changes the number of parameters of the model slightly, but we keep the same model names, like 816m, even though it now has 764m non-embedding parameters.

We use a batch size of 192 and sequence length 512 for our experiments, unless stated otherwise.

For RoPE we use the defaults from torchtune, which sets \texttt{base} to 10,000.
When applied to characters, the sequence length is effectively very short, so that only the first 64 to 128 dimensions of the character embeddings are affected.

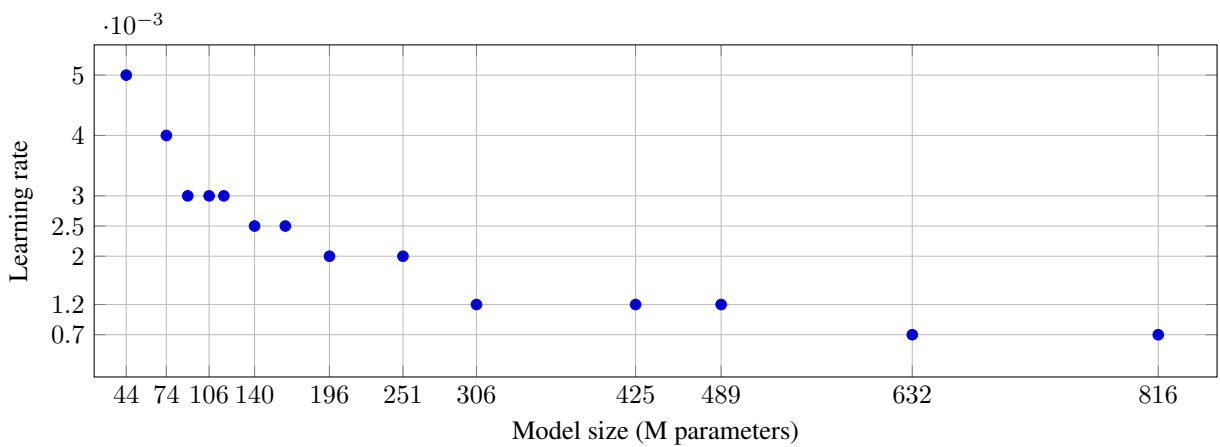
\begin{figure}
\centering
\begin{tikzpicture}
\begin{axis}[
    width=\textwidth,
    height=6cm,
    xlabel={Model size (M parameters)},
    ylabel={Learning rate},
    xmin=20,
    xmax=860,
    ymin=0,
    ymax=0.0055,
    xtick={0,44,74,106,140,196,251,306,425,489,632,816},
    ytick={0.0007,0.0012,0.002,0.0025,0.003,0.004,0.005},
    grid=both,
    enlargelimits=false,
]

\addplot+[only marks, mark=*] coordinates {
    (44, 0.005)
    (74, 0.004)
    (90, 0.003)
    (106, 0.003)
    (117, 0.003)
    (140, 0.0025)
    (163, 0.0025)
    (196, 0.002)
    (251, 0.002)
    (306, 0.0012)
    (425, 0.0012)
    (489, 0.0012)
    (632, 0.0007)
    (816, 0.0007)
};
\end{axis}
\end{tikzpicture}
\caption{Optimal learning rate as function of model size. Model sizes indicate the configurations listed in the Chinchilla paper~\citep{hoffmann2022training}. Actual parameter counts deviate due to changes in vocabulary size and the use of GQA.}
\label{fig:learning-rate}
\end{figure}

\paragraph{Chinchilla optimal training.}
This paper studies the tradeoff between training FLOPs and model quality.
So we adopt the 20:1 ratio of tokens to parameters found to be optimal by~\citet{hoffmann2022training}.
We validated on the base model, that this ratio is roughly optimal for our model and the Slimpajama dataset.

Today, it is customary to train models far beyond the Chinchilla-optimal point.
This makes sense when the inference cost dominates the training cost of the model.
However, the Chinchilla-optimal training ratio is still relevant when producing models for distillation.

\paragraph{Optimizer.} We use the AdamW optimizer with $\epsilon=10^{-7}$, $\beta_1=0.9$, $\beta_2=0.995$, and weight decay $0.1$.

We adjust the learning rate according to the size of the model, as depicted in Fig.~\ref{fig:learning-rate}
These were determined by training a model for each model size with 3 to 5 different learning rates and choosing the learning rate that achieved the best loss on the validation set.
We then used the same learning rates for the models with spelling bee embeddings.

\begin{table}[b]
    \centering
    \begin{tabular}{lll}
        Weight type & Variance at initialization\\\hline
        Token embeddings & $1/\sqrt{\text{embedding\_dim}}$ \\
        Character embeddings & $1/\sqrt{\text{embedding\_dim}}$ \\
        SwiGLU up projections & $1/\sqrt{\text{fan\_in}}$ * 1.679 & \citet{yang2024kolmogorov} \\
        Layer norm weights & 1.0\\
        All other weights & $1/\sqrt{\text{fan\_in}}$
    \end{tabular}
    \vspace{.3cm}
    
    \caption{Initialization parameters used for our experiments.}
    \label{tab:init}
\end{table}

\paragraph{Learning rate schedule.}
We use a linear warmup phase of 500 steps for all our experiments.
We have carefully compared shorter and longer warmup phases across different model sizes and found that 500 warmup steps work well across the model sizes considered in this work.

After the warmup phase, we use linear decay from the maximum learning rate to 10\% of the maximum learning rate.

\paragraph{Initialization.}
We experimented with the base models to find the best initialization and then used the same initialization for all our experiments.
All weights are initialized with a zero-centered normal distribution with different variances.
Table~\ref{tab:init} lists the variances for the different kinds of weights in the model.
An important insight was the need for the scaling factor for the SwiGLU up projections discussed by~\cite{yang2024kolmogorov}.

\begin{figure}[t]
    \centering
    \includegraphics[width=0.7\linewidth]{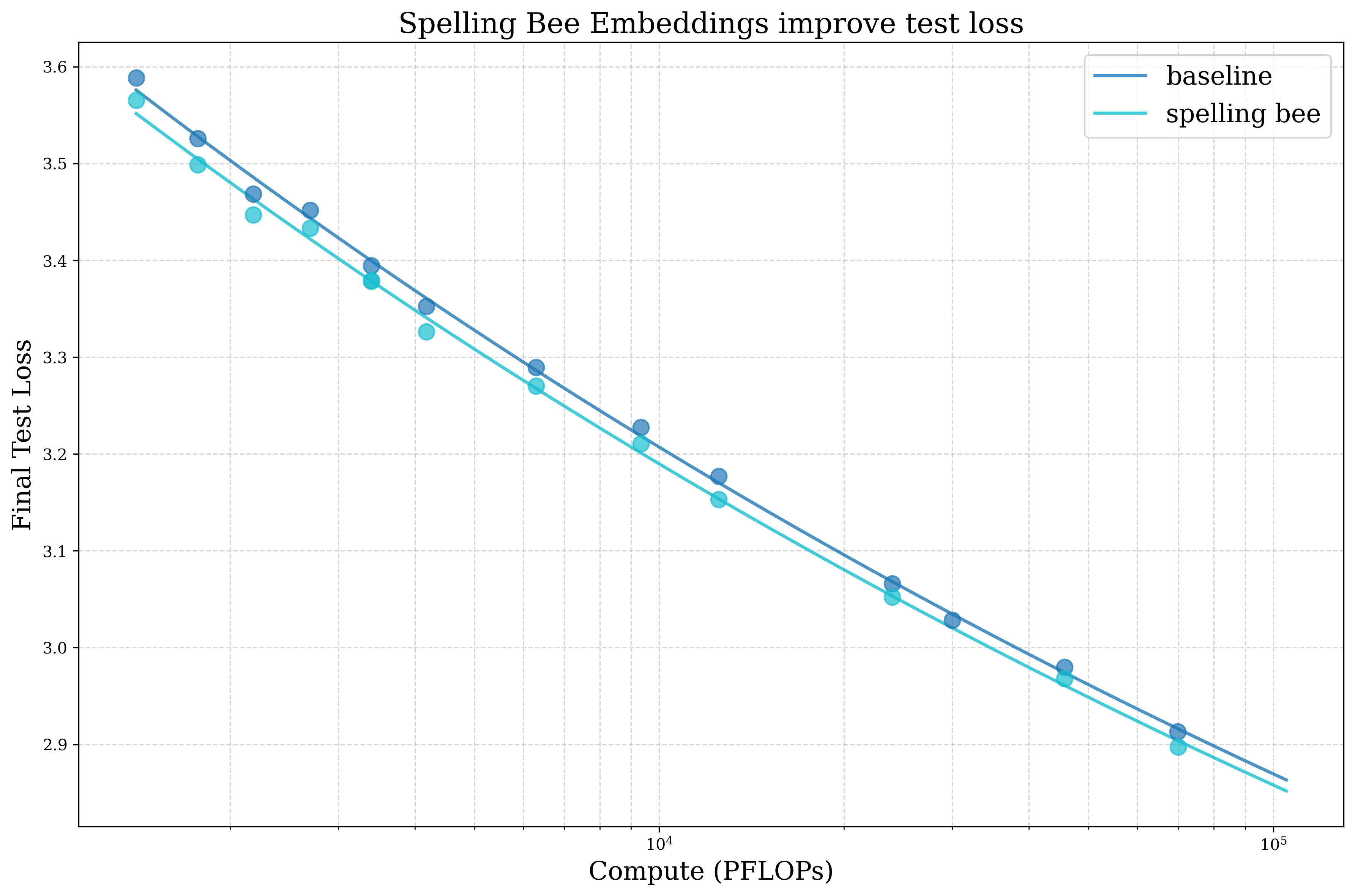}
    \caption{Scaling study using the test loss. We trained a series of models with growing parameter counts, with and without spelling bee embeddings.
    Each dot represents a different training run according to the methodology described in Section~\ref{sec:setup}.
    The lines are fitted to the data points with a shifted power law.
    }
    \label{fig:scaling}
\end{figure}

\paragraph{Counting FLOPs.}
To discuss compute-normalized results, we need a reliable way to measure FLOPs.
In line with other scaling studies, we estimate FLOPs to be $6pn$, where $p$ is the non-embedding parameters\footnote{This includes the final projection of the transformer, but not the initial embedding table} and $n$ is the number of tokens processed.
This calculation ignores all normalization steps, residual addition operations, the cost of attention, and the final loss computation.
Of these the attention FLOPs are the largest contributor.
However, for the short sequence lengths (512) that we consider in this work, their contribution is still negligible:
For the 44m model, the matmuls (sans attention, but including the final projection) require around 4.5e+11 FLOPs for the forward+backward pass of a single sequence.
In contrast, the attention operation requires around 1.3e+10 FLOPs, i.e. about 3\%.
For larger models the relative contribution of attention only drops further.

\section{Results}
\label{sec:results}

Spelling bee embeddings can help with spelling problems (Section~\ref{ssect:strawberry}), but our primary interest is their impact on established benchmarks.
We use two main ways to compare models: the loss on the SlimPajama test set and the scores determined by the \texttt{lm-evaluation-harness}~\citep{eval-harness}.

\begin{table}[t]
    \centering
    \begin{tabular}{l|l|c|c|c}
    \toprule
         & \textbf{Benchmark/Metric} & n-shot & Transformer & +Spelling Bee Embeddings \\ \specialrule{1.2pt}{1pt}{1pt}
         & Total parameters & & 918m & 918m \\
         & Non-emb. parameters & & 764m & 764m \\ \midrule
        Test Loss & SlimPajama & &  2.912 & \textbf{2.897} \\ \midrule
        English & AGIEval & 0 & 0.2585 & \textbf{0.2590} \\
        & ARC (easy) & 5 & 0.5173 & \textbf{0.5387} \\
        & ARC (challenge) & 5 & 0.2670 & \textbf{0.2764} \\
        & BBH & 3 & 0.0180 & \textbf{0.0237} \\
        & DROP (f1) & 0 & 0.0382 & \textbf{0.0490} \\
        & HellaSwag & 0 & 0.4204 & \textbf{0.4247} \\
        & MMLU & 0 & \textbf{0.2331} & 0.2302 \\
        & NaturalQuestions (open) & 0 & 0.0166 & \textbf{0.0169} \\
        & OpenBookQA & 0 & 0.2960 & \textbf{0.3100} \\
        & PIQA & 0 & 0.6773 & \textbf{0.6839} \\ 
        & SciQ & 0 & 0.6760 & \textbf{0.7160} \\
        & TriviaQA & 0 & \textbf{0.0238} & 0.0160 \\
        & WinoGrande & 0 & 0.5295 & \textbf{0.5375} \\ \midrule
        Math & GSM8k & 5 & 0.0144 & 0.0121 \\
        & MBPP & 3 & 0.0000 & 0.0020 \\
        & MATH & 0 & 0.0008 & 0.0012 \\ \midrule
        Code & HumanEval & 0 & 0.0061 & 0.0122 \\ \midrule
        Non-English & C-Eval & 0 & 0.2303 & \textbf{0.2571} \\
        & KMMLU & 0 & 0.1305 & \textbf{0.2983} \\
        \bottomrule
    \end{tabular}
    \vspace{.3cm}
    \caption{Results on the lm-evaluation-harness with the "816m" configuration from the Chinchilla paper. Note the mismatch between the "816m" name (referring to the Chinchilla configuration table) and the actual number of parameters. This is due to the adoption of a larger vocabulary and the use of GQA.
    For multiple-choice benchmarks we use the length-normalized accuracy score (``acc\_norm'') as computed by the lm-eval package.
    }
    \label{tab:lmeval}
\end{table}

In Fig.~\ref{fig:scaling} we observe that spelling bee embeddings improve the test loss over the baseline.
As more compute typically improves models, the core argument is one of \textbf{compute efficiency}: comparing the fitted scaling laws along the horizontal lines suggests that spelling bee embeddings are comparable to spending about 8\% more FLOPs.

We also observe that the advantage is relatively constant across two orders of magnitude of compute.
This is encouraging for the potential of spelling bee embeddings to help improve much \textbf{larger models} as well.

In Table~\ref{tab:lmeval} we compare models that we trained with and without spelling bee embeddings on the commonly used lm-evaluation-harness~\citep{eval-harness}.
Most benchmarks improve with the addition of spelling bee embeddings.

\subsection{Counting Rs}
\label{ssect:strawberry}
As there is no established benchmark for spelling problems, we vibe checked the largest models with and without spelling bee embeddings on the famous strawberry question.
When prompted with ``\texttt{The number of times the letter R occurs in strawberry is }'', the baseline model answered with ``\texttt{2}'', while the model enhanced with spelling bee embeddings answered ``\texttt{3}''.

\subsection{Spelling Benchmark}
\label{ssect:spelling-benchmark}                                       To systematically evaluate the impact of spelling bee embeddings on character-level tasks, we introduce a benchmark consisting of three    
task types that probe different aspects of spelling knowledge:
\begin{itemize}
  \item \textbf{Count:} Given a word and a letter, count the occurrences of that letter.\\
  Format: \texttt{The number of times the letter A occurs in banana is }\,$\rightarrow$\,\texttt{3}
  \item \textbf{Index:} Given a word and a position, identify the letter at that position.\\
  Format: \texttt{Q: What is the third letter of the word 'banana'? A:}\,$\rightarrow$\,\texttt{n}
  \item \textbf{Reverse:} Given a word, produce its reversed spelling.\\
  Format: \texttt{cat reversed is }\,$\rightarrow$\,\texttt{tac}
\end{itemize}

\paragraph{Dataset construction.}
We sample words from common English words\footnote{\url{https://github.com/first20hours/google-10000-english}} and a comprehensive word list, filtering to words of 4--10 characters.
For each task, we provide three few-shot examples in the prompt.
We evaluate on 5,000 samples: 2,450 count, 2,450 index, and 100 reverse tasks.
Palindromes are excluded from the reverse task to prevent trivial solutions.

\paragraph{Evaluation.}
We use the \texttt{lm-evaluation-harness}~\citep{eval-harness} with greedy generation, stopping at the first newline. 
We report exact-match accuracy (case-insensitive) for the 816m model.

\begin{table}[t]
  \centering
  \begin{tabular}{l|c|c|c|c}
  \toprule
      \textbf{Task} & \textbf{n-shot} & \textbf{Baseline} & \textbf{+Spelling Bee} & \textbf{$\Delta$} \\ \midrule
      Count & 3 & 12.6\% & 27.4\% & \textbf{+14.8\%} \\
      Index & 3 & 10.3\% & 16.2\% & \textbf{+5.9\%} \\
      Reverse & 3 & 0.0\% & 0.0\% & +0.0\% \\ 
  \bottomrule
  \end{tabular}
  \vspace{.2cm}
  \caption{Results on the spelling benchmark for the 816m model.}
  \label{tab:spelling}
\end{table}

\paragraph{Results.}                                                   
Table~\ref{tab:spelling} presents the results.
The count task shows the largest improvement (+14.8\%), likely because counting requires aggregating information across all character positions---precisely what the sum of position-encoded byte embeddings provides.
The index task also improves substantially (+5.9\%), demonstrating that the model can leverage character position information encoded via RoPE in the spelling bee embeddings.

\paragraph{The reverse task.}
Interestingly, both models achieve 0\% accuracy on the reverse task.
Manual inspection reveals that models consistently output the \emph{forward} spelling (e.g., \texttt{wheat} instead of \texttt{taehw}).
This suggests that while spelling bee embeddings improve character recognition and counting, they do not enable sequential character manipulation.

The reversal operation requires algorithmic reasoning that goes beyond having access to character-level representations.                   
We include the reverse task with reduced sample size (100 samples) to document this limitation.

\subsection{Data Saturation}
In practice, models are often trained on far more data than the Chinchilla-optimal ratio of data and parameters would suggest.
To study whether the effects disappear after training models on more data, we trained models with the 106m configurations on 10$\times$ more data than the Chinchilla paper suggests, with and without spelling bee embeddings.
The experiments summarized in Table~\ref{tab:datasaturation} suggest that the advantage of spelling bee embeddings does not diminish with more~data.

\begin{table}[t]
    \centering
    \begin{tabular}{l|c|c||c|c}
    \toprule
        & Baseline & +Spelling bee & difference & compute adv. \\ \midrule
        1$\times$ Chinchilla & 3.392 & 3.367 & 0.025 & 9.8\% \\  
        10$\times$ Chinchilla &  3.172 & 3.151 & 0.021 & 12.5\% \\  
        \bottomrule
    \end{tabular}
    \vspace{.2cm}
    \caption{Test loss of models with the 106m configuration. In contrast to other experiments these models were trained with a sequence length of 1024. We define the \emph{compute advantage} to be the relative FLOPs savings of the spelling bee variant to reach the same quality as the baseline.}
    \label{tab:datasaturation}
\end{table}

\subsection{Ablations}

\begin{table}[t]
    \centering
    \begin{tabular}{c|c|c|c|c|c|c|c}
    \toprule
        Metric & Transformer & +Spelling Bee & Bias-only & No rotary & No tok. emb. & Shuffled & First char \\ \midrule
        Test Loss & 2.912 & \textbf{2.897} & 2.912 & 2.899 & 2.915 & 2.901 & 2.907 \\
        \bottomrule
    \end{tabular}
    \vspace{.3cm}
    \caption{Results of the ablation studies for which we used the "816m" configuration from the Chinchilla paper.}
    \label{tab:ablations}
\end{table}

To justify the design choices of spelling bee embeddings, we ran ablation studies that remove some of the components and show that this diminishes the gains we observed.
Table~\ref{tab:ablations} presents the numerical results.

\paragraph{Bias-only.} In this ablation we do not add spelling bee embeddings to the token embeddings, but instead add a single learned embedding to the token embedding.
This single learned embedding is shared between all token embeddings and acts as a \emph{bias} that could help rare tokens, which do not see many parameter updates otherwise.
In the ablation experiments with the "425m" and "816m" models, the bias-only option did not improve the test loss.
However, for smaller models we observed some gains.

\paragraph{No rotary.} To study the impact of rotational embeddings for the characters within the word, we remove the rotary embeddings.
The resulting model effectively presents the spelling of tokens as a ``bag of characters''.
The model appears to suffer only very little from this change, suggesting that the order of characters is not a major factor.
Although the difference is very small, it was consistent across many model sizes, so we believe it is beneficial to keep it, given the low compute overhead of rotary embeddings.

\paragraph{No token embedding.} Next, we investigate whether the embedding of the spelling itself is sufficient. So instead of averaging spelling embedding and token embedding (see Fig.~\ref{fig:spelling-bee-embeddings}), we only take the spelling embedding.
With a test loss of 2.915 vs 2.912 this seems to be ever so slightly slightly worse than even the baseline.
Given that this option reduces the number of parameters from 918m to 764m, this is a surprising result.
It also suggests that the token embeddings are not providing a lot of learning capacity.

\paragraph{Shuffled spelling table.} One way to understand the impact of spelling is to shuffle the spelling table, so that the spelling embeddings contain the spelling of a randomly chosen token.
We observe that this negatively impacts the test loss, but, surprisingly, it still improves over the baseline.

\paragraph{First character only.} Instead of giving the full spelling, we only provide the first character in this experiment. As expected, this reduces the benefit of spelling bee embeddings, but it still improves over the baseline.


\section{Limitations}
\label{sec:limitations}
The vast majority of architecture changes have ultimately not been adopted in modern large language models.
The most common failure mode of architecture research is that the baseline has not been tuned sufficiently.
We took great care in avoiding this effect by running over 1500 training runs to tune the hyperparameters, especially the learning rate, across multiple sizes of the baseline.
Further, we reproduced well-established papers, like the benefit of the SwiGLU activation function~\citep{shazeer2020glu}, to validate that our code produces observations consistent with the literature.

Our scaling studies go up to about 800m parameters.
However, the benefits may disappear at even larger scale.

The size of the vocabulary may affect the results.
We would expect that with much smaller vocabularies, e.g. for character-level vocabularies, the effects of spelling bee embeddings should disappear.

Given the right data, any modern language model architecture can learn the spelling of tokens.
It is possible that the effect we observe in this paper disappears for models that are trained on many more tokens, or with different datasets.
However, the fact that today's strongest language models are struggling with spelling problems today suggests that this is not the case.

Retrofitting spelling bee embeddings on models pretrained with regular embeddings might be possible, but hasn't been explored in this work.
This means that, at least for now, using spelling bee embeddings will need pretraining to be adopted.

Currently, the community has standardized around the Byte Pair Encoding (BPE)~\citep{gage1994new,sennrich2016neural} algorithm, but more optimal tokenization might alleviate the need for spelling-aware embeddings.

\section{Conclusion}
Infusing token embeddings with character-level structure leads to consistent improvements across model sizes and standard benchmarks; even though most of them are not obviously related to spelling challenges.
Besides improving existing language models, this opens the door to further improvements in tokenization algorithms and prompts us to revisit the established choices such as vocabulary size.

\paragraph{Acknowledgments}
This work was enabled by the generous support of Sutter Hill Ventures. This direction was inspired in part by DeLesley Hutchins' work at Google, who pioneered tokenization-free models.
We also thank Cyris Kissane for an elegant explanation for preliminary results regarding the ``bias-only'' ablation study.

\bibliographystyle{unsrtnat}
\bibliography{references}  

\newpage
\appendix
\section*{Appendix}
\section{Spelling still is a challenge: Examples}
\label{app:screenshots}
Two years after the issue of counting the Rs in strawberry has been discussed first, the most advanced language models do answer the question correctly when asked directly.
However, even slight variations still can trip up these models.
The following screenshots were taken on Nov 16, 2025.
These examples were not hard to find and are just a selection of a few attempts made to trip up the models.

With ChatGPT 5:
\begin{center}
\includegraphics[width=12cm]{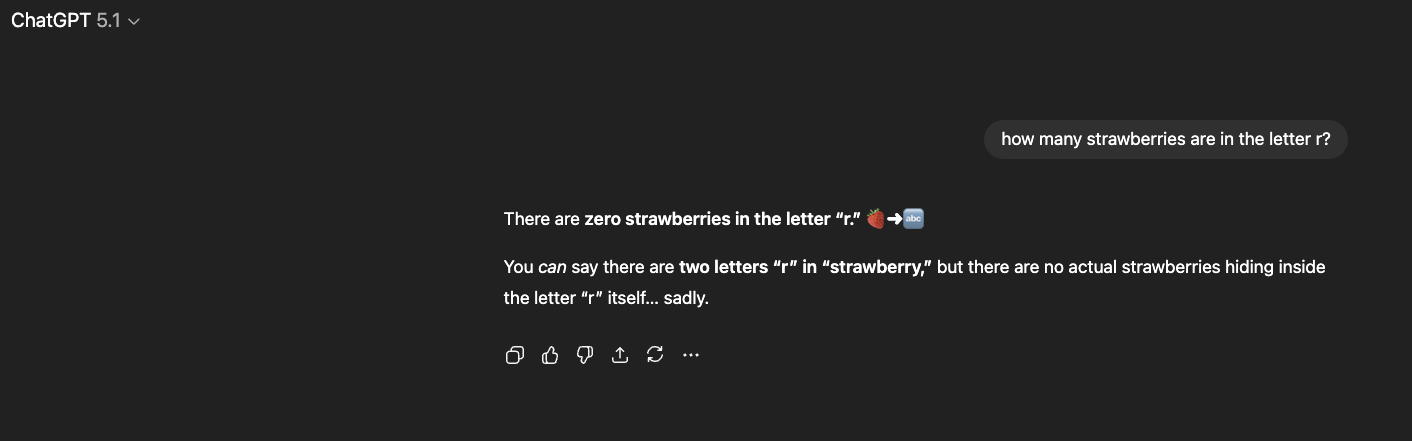}
\end{center}

Same with Claude 4.5:
\begin{center}
\includegraphics[width=12cm]{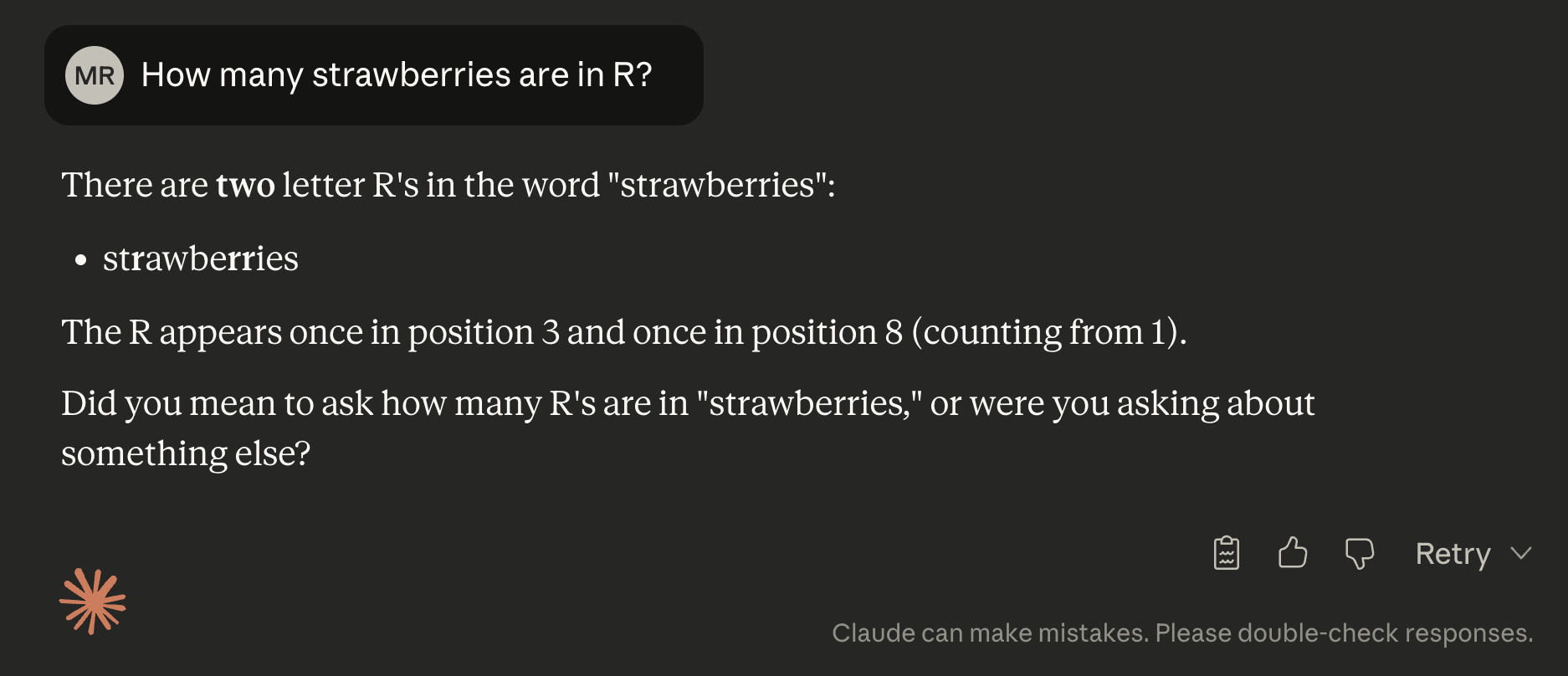}
\end{center}

Variant with Claude 4.5
\begin{center}
\includegraphics[width=12cm]{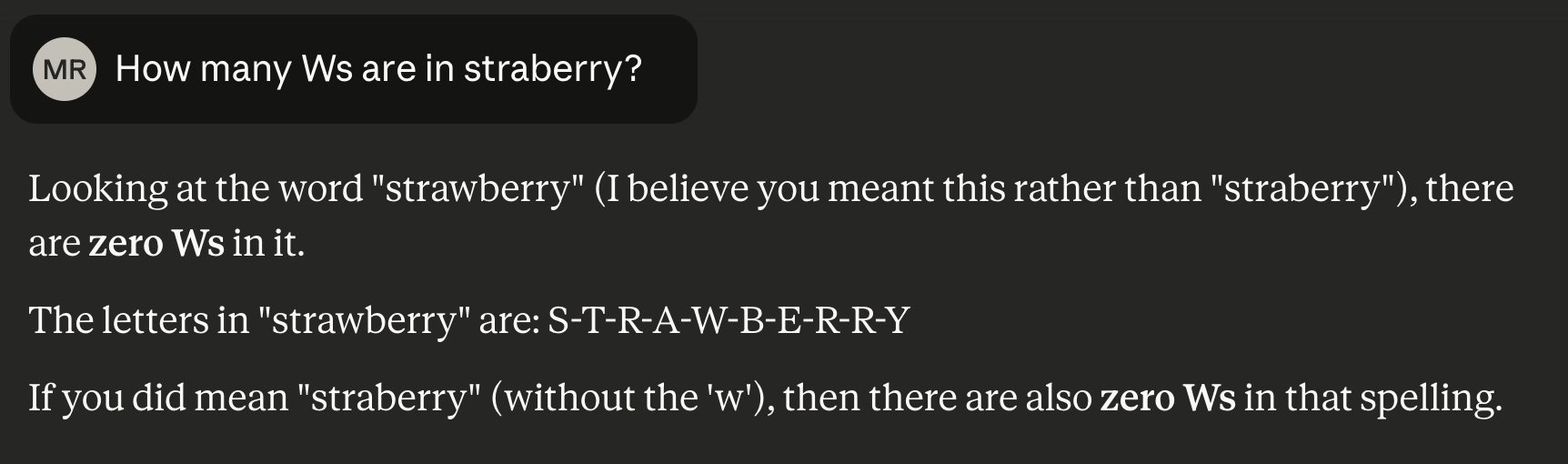}
\end{center}

Misspelling `strawberry` with an extra R and adding a parenthesis as a confusion is enough to trip up Gemini.
\begin{center}
\includegraphics[width=12cm]{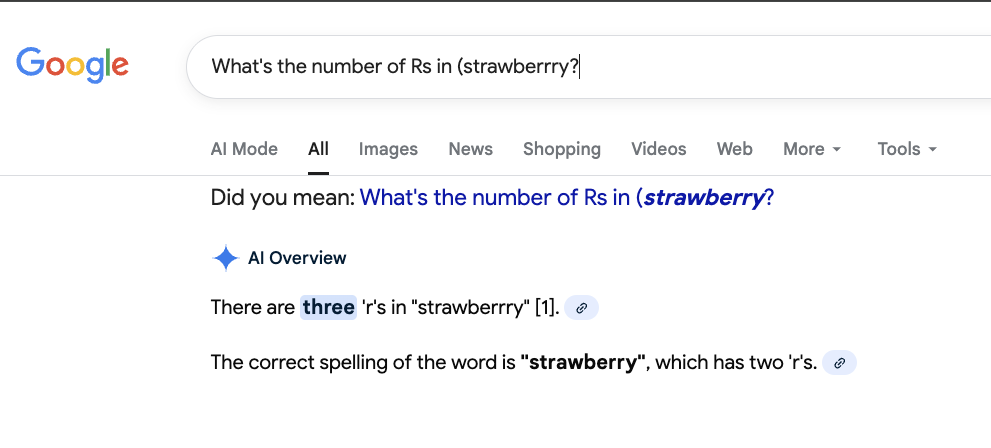}
\end{center}

Counting Rs in a made-up word:
\begin{center}
\includegraphics[width=12cm]{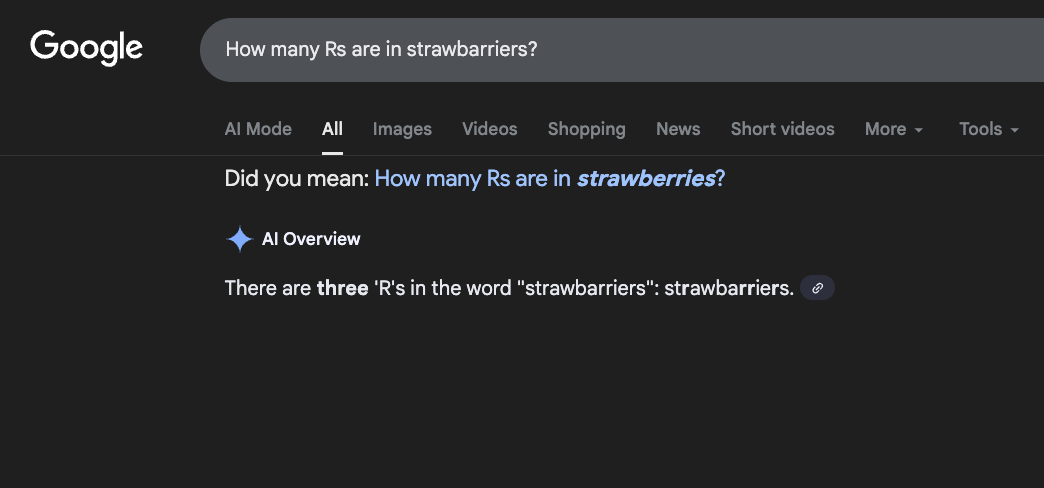}
\end{center}

Misspelled words are hard:
\begin{center}
\includegraphics[width=12cm]{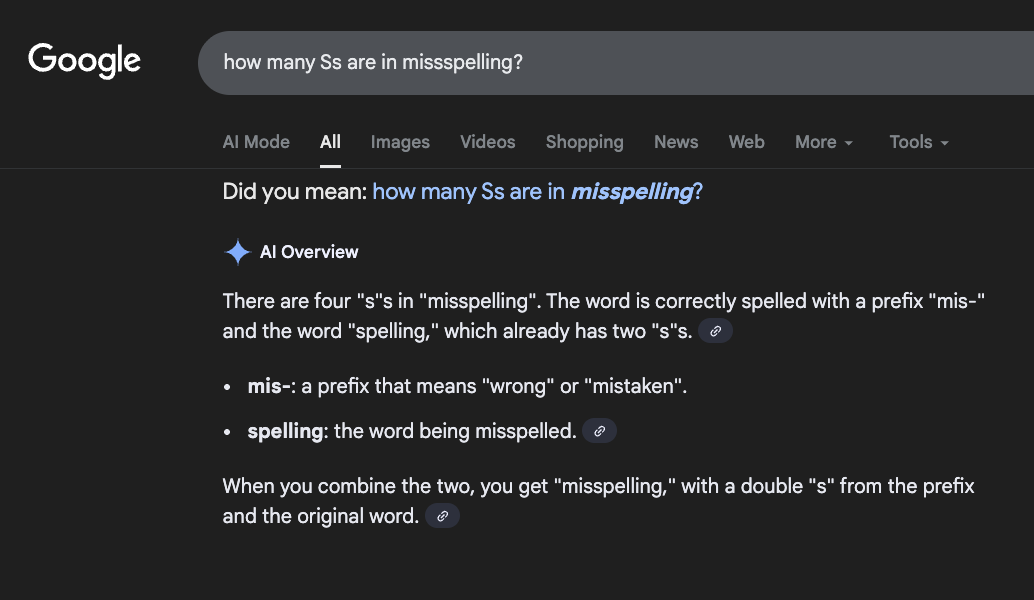}
\end{center}

\section{Are models explicitly trained on the strawberry question?}
Evidence that the question of counting Rs in strawberry has been explicitly trained into recent models. Screenshots taken on Jan 7, 2026:
\begin{center}
\includegraphics[width=12cm]{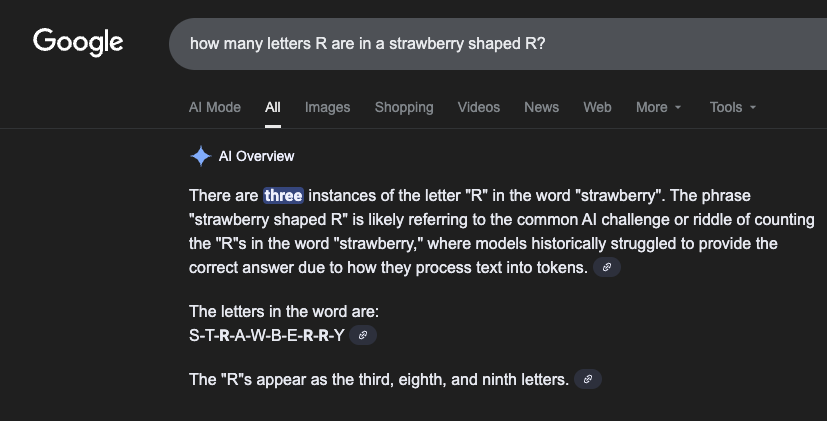}
\end{center}
\begin{center}
\includegraphics[width=12cm]{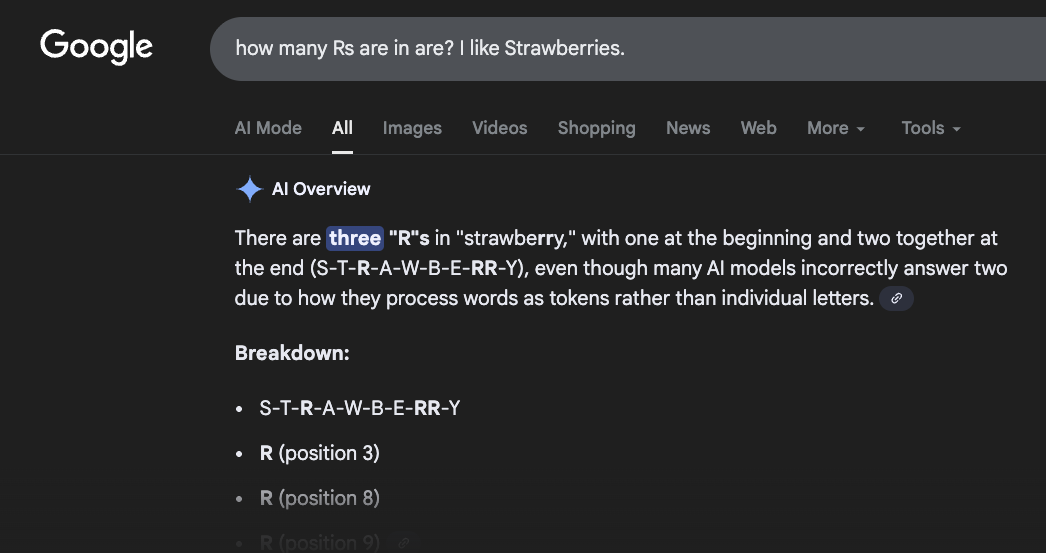}
\end{center}

\section{Reproduced Studies}
As a sanity check, we studied the results by \citet{shazeer2020glu} 
and were able to reproduce them.

Here are scaling plots for the comparison of different activation functions:
\begin{center}
\includegraphics[width=0.8\textwidth]{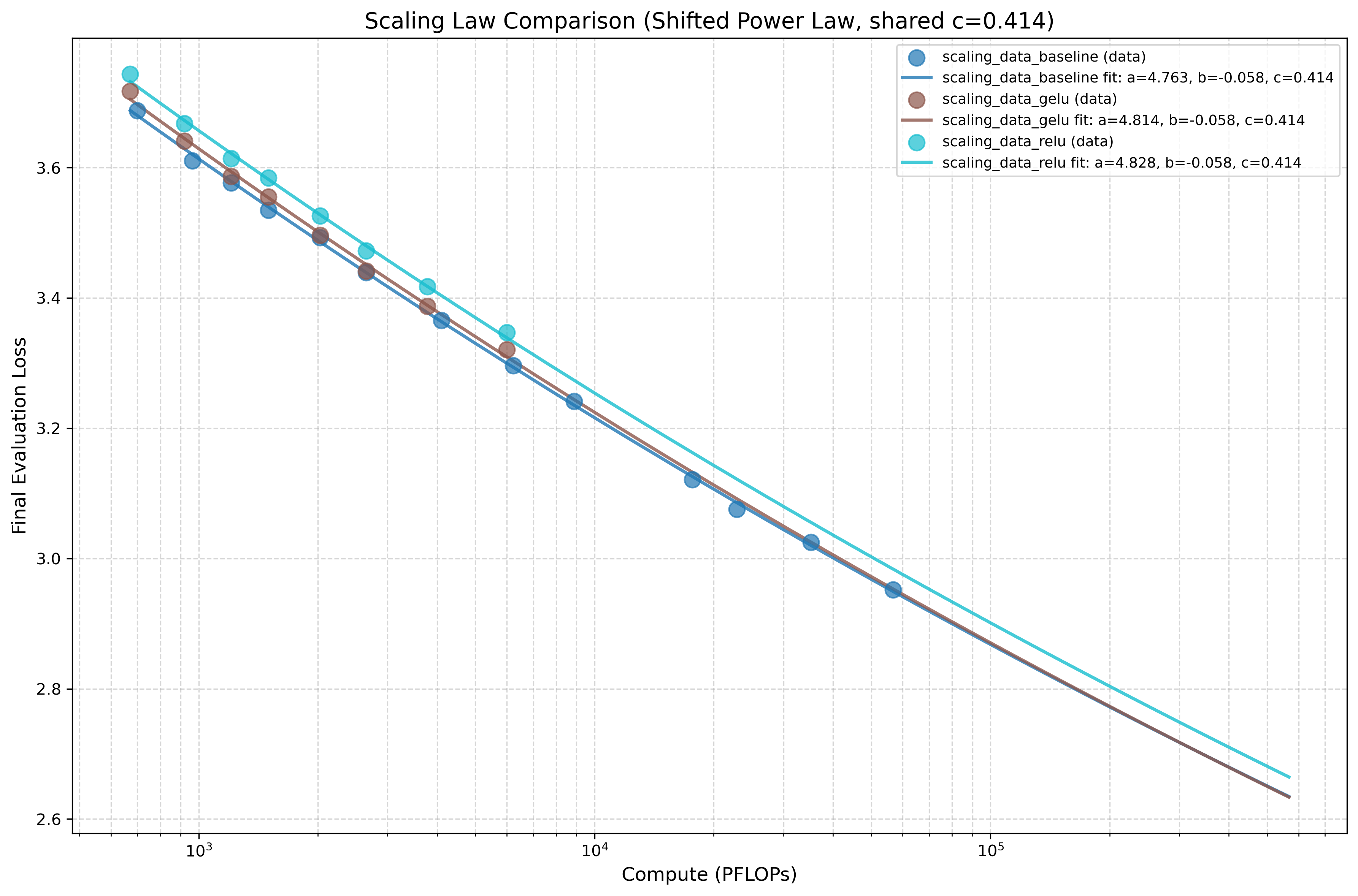}
\end{center}

\end{document}